%% file: main.tex
\documentclass[acmtog]{acmart}

\AtBeginDocument{%
  }

\usepackage{multirow}
\usepackage{array}
\usepackage{pifont}
\newcommand*\colourcheck[1]{%
  \expandafter\newcommand\csname #1check\endcsname{\textcolor{#1}{\ding{52}}}%
}
\newcommand*\colourmark[1]{%
  \expandafter\newcommand\csname #1mark\endcsname{\textcolor{#1}{\ding{55}}}%
}
\newcommand{\boldstartspace}[1]{\vspace{0.05in}\noindent\textbf{#1}}
\newcommand{\revise}[1]{\textcolor{black}{#1}}
\colourcheck{blue}
\colourcheck{green}
\colourcheck{red}
\colourcheck{black}

\colourmark{blue}
\colourmark{green}
\colourmark{red}
\colourmark{black}

\usepackage{xcolor} 
\PassOptionsToPackage{pagebackref,breaklinks,colorlinks,citecolor=citeblue,bookmarks=false,linkcolor=blue,urlcolor=blue}{hyperref}
\usepackage{hyperref}
\usepackage[capitalize]{cleveref}  
\crefname{section}{Sec.}{Secs.}
\Crefname{section}{Section}{Sections}
\crefname{table}{Tab.}{Tabs.}
\Crefname{table}{Table}{Tables}
\crefname{figure}{Fig.}{Figs.}
\Crefname{figure}{Figure}{Figures}
\crefname{equation}{Eq.}{Eqs.}
\Crefname{equation}{Equation}{Equations}

\newcommand{\tocite}[1]{\textcolor{red}{[TO CITE]}}
\newcommand{\toref}[1]{\textcolor{red}{[TO REF]}}

\input{symbols_format.tex}

\acmSubmissionID{181}

\begin{document}

\title[RigAnything]{RigAnything: Template-Free Autoregressive Rigging for Diverse 3D Assets}

\author{Isabella Liu}
\email{lal005@ucsd.edu}
\affiliation{%
  \institution{UC San Diego}
  \country{USA}
}

\author{Zhan Xu}
\email{zhaxu@adobe.com}
\affiliation{%
  \institution{Adobe Research}
  \country{USA}
}

\author{Wang Yifan}
\email{yifwang@adobe.com}
\affiliation{%
  \institution{Adobe Research}
  \country{USA}
}

\author{Hao Tan}
\email{hatan@adobe.com}
\affiliation{%
  \institution{Adobe Research}
  \country{USA}
}

\author{Zexiang Xu}
\email{zexiangxu@gmail.com}
\affiliation{%
  \institution{Hillbot Inc.}
  \country{USA}
}

\author{Xiaolong Wang}
\email{xiw012@ucsd.edu}
\affiliation{%
  \institution{UC San Diego}
  \country{USA}
}

\author{Hao Su}
\email{haosu@ucsd.edu}
\affiliation{%
  \institution{UC San Diego}
  \country{USA}
}
\affiliation{%
  \institution{Hillbot Inc.}
  \country{USA}
}

\author{Zifan Shi}
\email{vivianszf9@gmail.com}
\affiliation{%
  \institution{Adobe Research}
  \country{USA}
}
\renewcommand{\shortauthors}{Liu et al.}

\begin{abstract}
We present \textbf{\emph{RigAnything}}, a novel autoregressive transformer-based model, which makes 3D assets rig-ready by probabilistically generating joints and skeleton topologies and assigning skinning weights in a template-free manner.
Unlike most existing auto-rigging methods, which rely on predefined skeleton templates and are limited to specific categories like humanoid, RigAnything approaches the rigging problem in an autoregressive manner, iteratively predicting the next joint based on the global input shape and the previous prediction.
While autoregressive models are typically used to generate sequential data, RigAnything extends its application to effectively learn and represent skeletons, which are inherently tree structures. To achieve this, we organize the joints in a breadth-first search (BFS) order, enabling the skeleton to be defined as a sequence of 3D locations and the parent index.
Furthermore, our model improves the accuracy of position prediction by leveraging diffusion modeling, ensuring precise and consistent placement of joints within the hierarchy. This formulation allows the autoregressive model to efficiently capture both spatial and hierarchical relationships within the skeleton.
Trained end-to-end on both RigNet and Objaverse datasets, RigAnything demonstrates state-of-the-art performance across diverse object types, including humanoids, quadrupeds, marine creatures, insects, and many more, surpassing prior methods in quality, robustness, generalizability, and efficiency. 
\revise{It achieves significantly faster performance than existing auto-rigging methods, completing rigging in under a few seconds per shape.}
Please check our website for more details: \href{https://www.liuisabella.com/RigAnything}{\textcolor{blue}{\textsf{https://www.liuisabella.com/RigAnything}}}.
\end{abstract}

\setcopyright{acmlicensed}
\acmJournal{TOG}
\acmYear{2025} \acmVolume{44} \acmNumber{4} \acmArticle{} \acmMonth{8}\acmDOI{10.1145/3731149}




\begin{teaserfigure}
    \includegraphics[width=\textwidth]{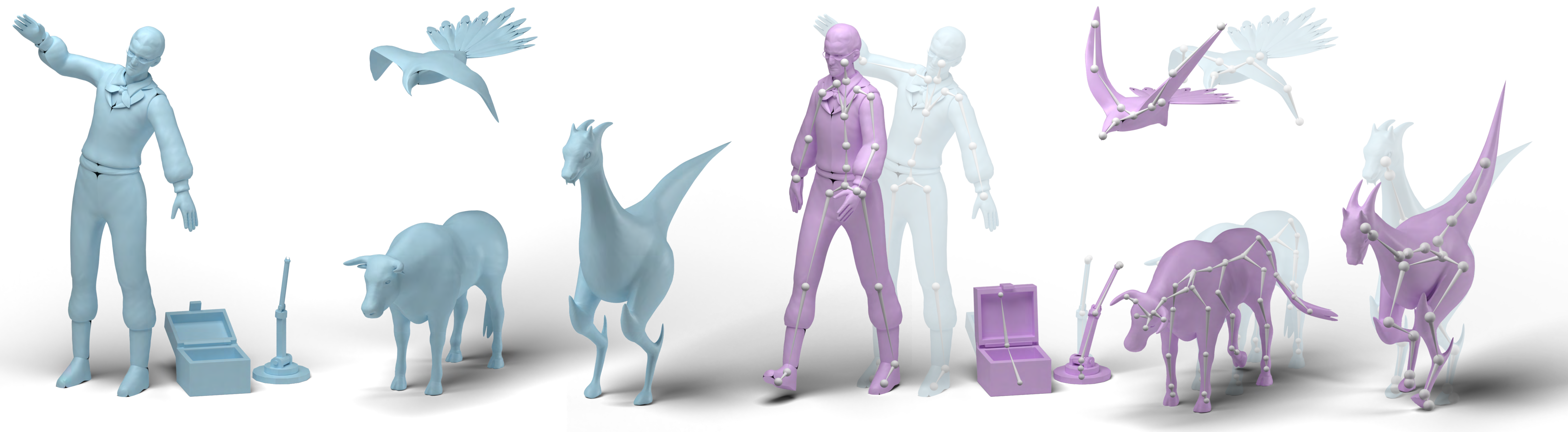}
    \caption{RigAnything is an autoregressive transformer-based approach for automatic rigging. From an arbitrarily posed shape (shown on the left), it can generate a skeleton and skinning weights that adapt seamlessly to the input's global structure (shown on the right), enabling articulation into new poses.
    }
    \label{fig:teaser}
\end{teaserfigure}
\maketitle

\input{Sections/1_introduction}

\input{Sections/2_related_work}

\input{Sections/3_method}

\input{Sections/4_experiments}

\input{Sections/5_discussion}

\input{Sections/6_conclusion_future_works}

\bibliographystyle{ACM-Reference-Format}
\bibliography{reference}

\end{document}

%% file: symbols_format.tex
\usepackage{xcolor}

\newcommand{\Shape}{\mathcal{S}}

%% file: Sections/1_introduction.tex
\section{Introduction}

\begin{figure*}[t]
\includegraphics[width=\linewidth]{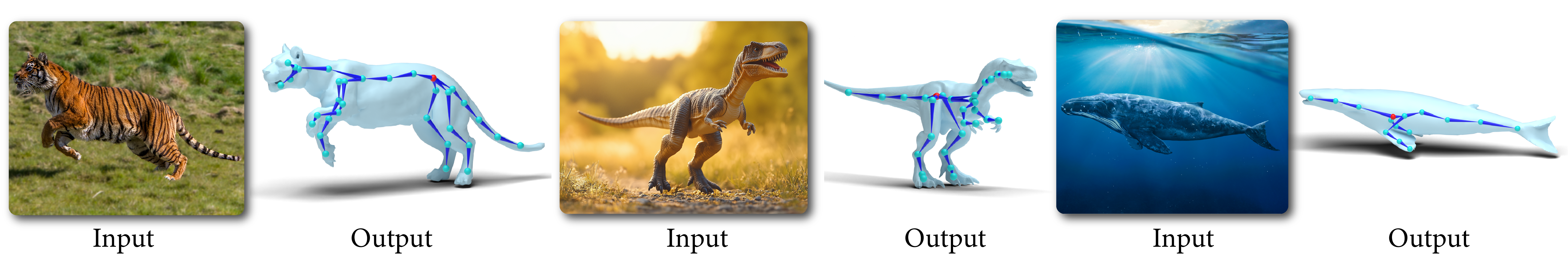}
\captionsetup{labelfont={color=black},textfont={color=black}}
\caption{Skeleton generation given real images, showing our method generalizes well to real data. More real results in \cref{fig:real_capture}.}
\label{fig:real_teaser}
\end{figure*}

Recent advancements in large-scale 3D asset generation~\cite{hong2023lrm, li2023instant3d,xu2023dmv3d,jun2023shap, shi2023mvdream,nichol2022point,liu2023zero,liu2024one} have enabled the creation of highly detailed static shapes. 
However, since motion is an essential aspect of how humans perceive and interact with the world, there is a growing demand for modeling dynamics to create lifelike and interactive assets \cite{liu2024dynamic}.
While some approaches leverage text-based~\cite{bahmani20244d, zhao2023animate124,singer2023text} or video-guided~\cite{yin20234dgen,ren2023dreamgaussian4d} control to animate objects, these methods often fall short in providing the precision and flexibility required by artists to fully realize their creative visions.
Rigging, in contrast, offer a robust and artist-friendly framework for animation, enabling fine-grained control over degree of freedom and range of motion. Our work addresses this need by presenting a systematic approach to automating rigging, advancing the state of the art in articulable asset generation. 

Auto-rigging has long been a challenging research problem in computer graphics~\cite{pinocchio, rignet, guo2024makeitani, chu2024humanrig, li2021learning}. \Cref{tab:teaser_comp} provides a concise summary of state-of-the-art methods in this domain.
Most existing approaches depend on predefined skeleton templates~\cite{pinocchio, guo2024makeitani, chu2024humanrig, li2021learning}, which limit their applicability to specific categories, such as humanoid characters (\cref{tab:teaser_comp}).
To overcome template reliance, RigNet~\cite{rignet} employs non-differentiable operators, including clustering for joint position acquisition and a minimum spanning tree for topology construction. However, this approach requires approximately two minutes to rig a single object and is further constrained to operate only on objects in rest poses.

In this work, we propose a transformer-based autoregressive model, termed \emph{\textbf{RigAnything}}, to make any 3D asset "rig-ready". The autoregressive model probabilistically "grows" the skeleton from the root joint in a sequential manner; Skinning weights for any surface sample are then inferred by holistically considering all the joints.

\input{Tabels/teaser_tab}

Specifically, we represent the tree-structured skeleton as a sequence by ordering the joints in a breadth-first search (BFS) order, where each joint is defined by a 3D position and a parent index. 
This autoregressive formulation is particularly suited for skeleton prediction, as it addresses the inherent ambiguity in joint configurations by representing them as a probabilistic distribution. Additionally, by sequentially generating joints and connections without relying on a predefined template, the model supports arbitrary skeleton structures and varying numbers of joints, enabling broad generalization across diverse object categories.
Furthermore, while \revise{transformer-based} autoregressive models are traditionally designed to handle discrete values~\cite{waswani2017attention,brown2020language,radford2019language}, inspired by recent work utilizing autoregressive models for image generation~\cite{li2024autoregressive}, we adopt a diffusion sampling process to predict the continuously valued joint positions, resulting in superior accuracy.
Given the predicted skeleton, we infer the skinning weights by a pair-wise computation.
We employ transformer blocks throughout the model to comprehensively capture the global shape structure, as well as the interdependence among all joints and their associated surface points.

We train our model end-to-end on both the RigNet dataset~\cite{rignet} and a curated subset of high-quality animatable assets from the Objaverse dataset~\cite{deitke2023objaverse}. We rigorously filter the Objaverse dataset and select 9686 high-quality rigged shapes, which enrich the dataset for research in this direction. The input shapes are further augmented with random pose variations to enhance robustness.
Our training data encompasses a wide range of object types, including bipedal, quadrupedal, avian, marine, insectoid, and manipulable rigid objects, as well as a diverse set of initial poses. This extensive scale and diversity of training data surpasses all prior work, playing a critical role in achieving broad generalizability across shape categories and configurations.

Extensive experiments demonstrate that RigAnything achieves state-of-the-art performance in the auto-rigging task, as demonstrated in \cref{fig:teaser}, \cref{fig:real_teaser} and \cref{fig: skeleton_comp}, surpassing prior methods in quality, robustness, generalizability, and efficiency. By automating rigging for diverse 3D assets, our method advances the vision of fully interactive 3D environments and scalable 3D content creation, empowering artists and developers with a powerful, efficient tool.

%% file: Tabels/teaser_tab.tex
\begin{table}[t]
    \small
    \centering
    \setlength{\tabcolsep}{1pt}
    \scalebox{0.75}{
    {\fontsize{8pt}{10pt}\selectfont
    \begin{tabular}{ccccccc}
    \toprule
    \multicolumn{1}{c}{\textbf{Methods}} & \multicolumn{1}{c}{\textbf{Humanoid}} & \multicolumn{1}{c}{\begin{tabular}[c]{@{}c@{}}\textbf{Non-}\\ \textbf{Humanoid}\end{tabular}} &\multicolumn{1}{c}{\begin{tabular}[c]{@{}c@{}}\textbf{Template-}\\ \textbf{Free}\end{tabular}}& \multicolumn{1}{c}{\begin{tabular}[c]{@{}c@{}}\textbf{Arbitrary}\\ \textbf{Pose}\end{tabular}} & \multicolumn{1}{c}{\begin{tabular}[c]{@{}c@{}}\textbf{Rigging}\\ \textbf{Time} \revise{(A100)}\end{tabular}} \\ \toprule
    TARig~\cite{ma2023tarig} & \greencheck & \redmark & \redmark & \redmark & $\sim$40s \\ 
    Pinocchio~\cite{pinocchio} & \greencheck & \greencheck & \redmark & \redmark & $\sim$40s  \\ 
    RigNet~\cite{rignet} & \greencheck & \greencheck  & \greencheck & \redmark&$\sim$120s \\
    \textbf{RigAnything (Ours)} & \greencheck & \greencheck & \greencheck & \greencheck & $\sim$2s \\ \bottomrule  
    \end{tabular}}}
    \caption{Feature comparison with other Auto-Rigging tools.}
    \label{tab:teaser_comp}
\end{table}

%% file: Sections/2_related_work.tex
\section{Related Work}

\begin{figure*}[t]
  \centering
  \includegraphics[width=\linewidth]{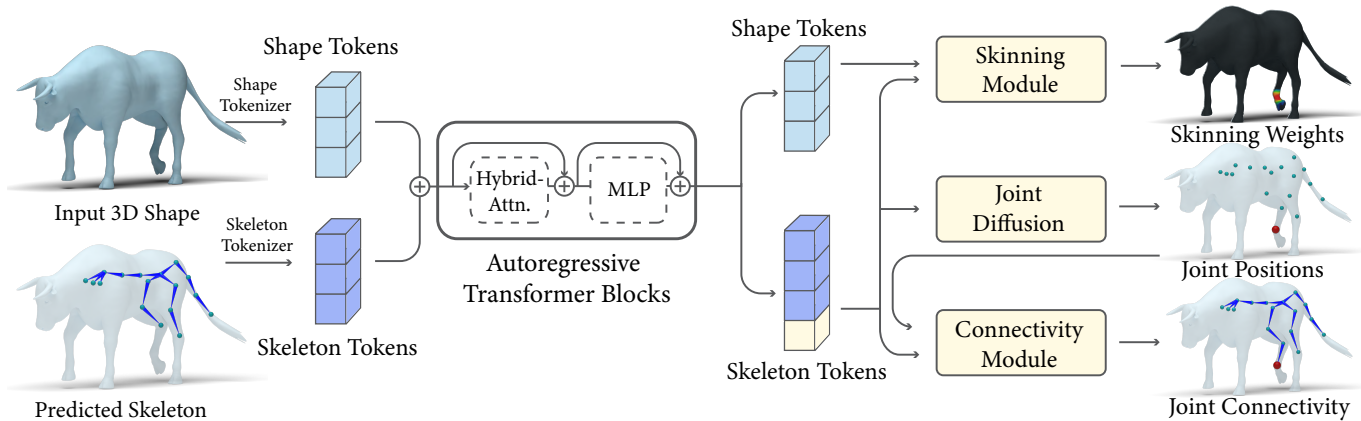}
  \caption{\revise{A single step in our method:} The input shape and the previously predicted skeleton sequence are tokenized using two separate tokenizers. These tokens are processed through a chain of autoregressive transformer blocks with a hybrid attention mask. Shape tokens perform self-attention to capture global geometric information, while skeleton tokens attend to all shape tokens and use causal attention within themselves to maintain the autoregressive generation process. After the transformer blocks, a skinning module decodes shape tokens into skinning weights, a joint diffusion module samples the next joint position, and a connectivity module predicts the next joint's connection to its preceding joints.}
  \label{fig:main_pipeline}
\end{figure*}

\subsection{Automatic Rigging}

Rigging is a fundamental technique for animation in computer graphics.
Traditional automatic rigging methods, such as Pinocchio~\cite{pinocchio}, rely on predefined skeletons and optimize their variations to fit a range of characters, with skinning weights determined by analyzing vertex-bone deformation relationships.
However, the optimization process is computationally expensive and diminishes the generalizability.
Recent advances in deep learning have improved the quality and adaptability of rigging.
TARig~\cite{ma2023tarig} utilizes a template with adaptive joints and a boneflow field to generate skeletons and skinning weights for humanoid characters.
\revise{Li’s et al~\shortcite{li2021learning}} leverages a predefined skeleton template for characters to learn rigging and proposes neural blend shape to enhance deformation quality.
However, these methods are confined to humanoid characters in standard poses and rely heavily on predefined templates, limiting their robustness and generalization to diverse objects, poses, and skeleton topologies.

Differently, RigNet~\cite{rignet} use a combination of regression and adaptive clustering to handle the diverse number of joints and employs a deep neural network for connectivity prediction to allow various topologies without templates or assumptions about shape classes and structures.
However, it lacks the robustness and efficiency due to its model design, which is not end-to-end trainable with clustering and Minimum Spanning Tree operations.
Make-it-Animatable~\cite{guo2024makeitani} and HumanRig~\cite{chu2024humanrig} are works developed concurrently with ours. They also focus solely on humanoid characters and rely on template skeletons, restricting their adaptability to more diverse data categories. 
In contrast, our method eliminates the need for templates and avoids assumptions about skeleton topology, achieving greater generalizability and robustness for diverse object types in a feed-forward manner.

\subsection{Autoregressive Models for 3D}

Autoregressive models are a powerful class of probabilistic models widely applied across domains such as natural language processing~\cite{brown2020language,achiam2023gpt,radford2019language} and computer vision~\cite{esser2021taming,parmar2018image,chen2020generative,li2024autoregressive}. 
In 3D tasks, autoregressive models have also demonstrated remarkable potential in areas like shape generation~\cite{yan2022shapeformer,mittal2022autosdf,ibing2023octree,cheng2022autoregressive,argus} and motion generation~\cite{t2mgpt,rempe2021humor,han2024amd}.
In 3D shape generation, methods mostly focus on designing effective representations for autoregressive modeling.
ShapeFormer~\cite{yan2022shapeformer} introduces a sparse representation that quantizes non-empty voxel grids in a predefined order.
AutoSDF~\cite{mittal2022autosdf} takes a different approach by modeling the entire space and using randomized sampling orders to enable non-sequential modeling.
Octree Transformer~\cite{ibing2023octree} introduce octree-based hierarchical shape representations with adaptive compression, significantly reducing sequence lengths.
Cheng \textit{et al.}~\cite{cheng2022autoregressive} decompose point clouds into semantically aligned sequences.
Argus3D~\cite{argus} utilizes discrete representation learning on a latent vector and scales up the model to improve the quality and versatility of 3D generation. 
Similarly, autoregressive models have advanced 3D motion generation.
\revise{MotionVAEs~\cite{ling2020character} uses autoregressive conditional variational autoencoders to learn a latent action space for human movement generation and control.}
T2M-GPT~\cite{t2mgpt} uses motion VQ-VAE and textural descriptions for human motion generation. 
HuMoR~\cite{rempe2021humor} proposes hierarchical latent variables for realistic motion synthesis.
AMD~\cite{han2024amd} presents an autoregressive model that iteratively generates complex 3D human motions from long textual descriptions.
In this paper, we pioneer the application of autoregressive models to the task of automatic rigging, marking a significant advancement in this domain.

%% file: Sections/3_method.tex
\section{Method}

Our goal is to transform a given 3D shape into an animatable asset by generating a plausible skeleton and predicting the corresponding skinning weights. These enable the 3D asset to be articulated under joint transformations using Linear Blend Skinning (LBS).
In this section, we first describe two types of skeleton ambiguities in \cref{sec:ambiguity}. We then present our novel autoregressive model for skeleton generation in \cref{sec:skeleton}, followed by our approach to skinning weight prediction in \cref{sec:skinning}. Finally, we provide a detailed description of the model architecture and the overall training objective in \cref{sec:transformer} and \cref{sec:final_loss}.

\subsection{Skeleton Ambiguity}\label{sec:ambiguity}
\begin{figure}[h]
\includegraphics[width=0.8\linewidth]{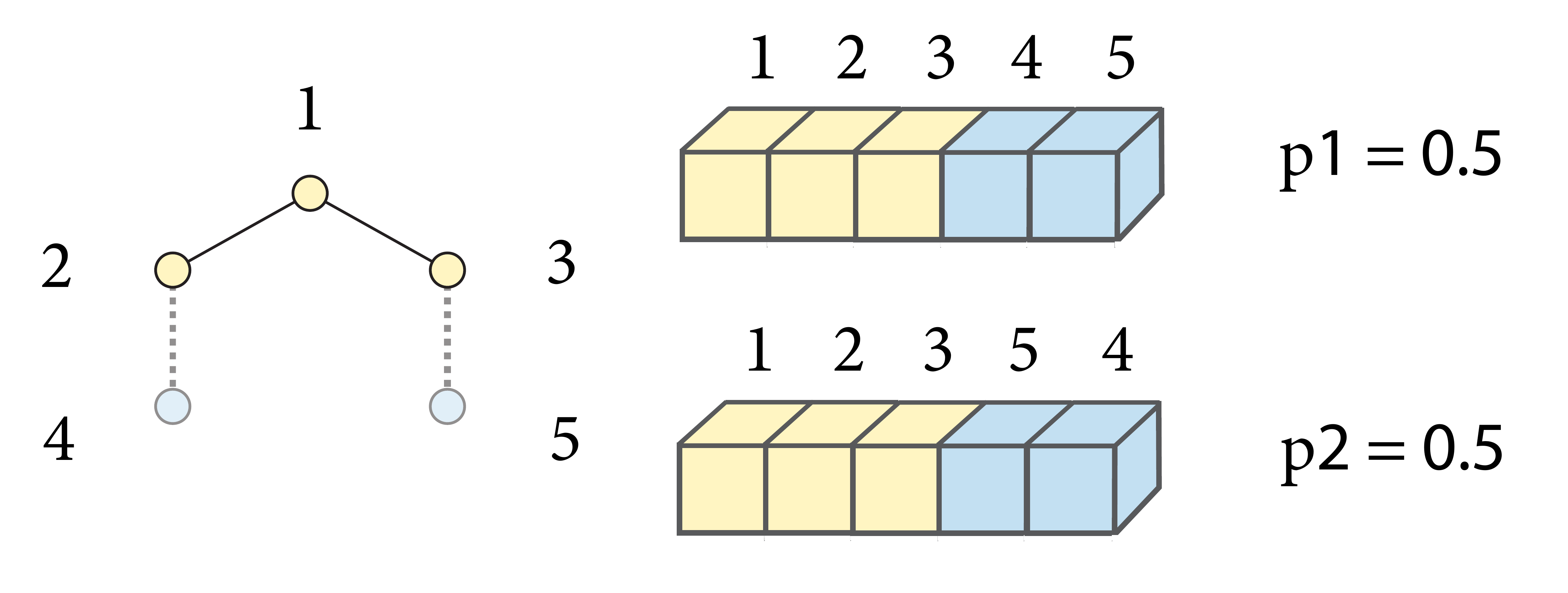}
\caption{Illustration of sibling ambiguity during BFS ordering in skeletons.}
\label{fig:sibling_ambiguity}
\end{figure}

We identify two types of skeleton ambiguities that commonly occur in existing pipelines.
\textbf{(1) Sibling ambiguity:} The ordering of nodes at the same depth in the skeleton tree is undefined, as illustrated in \cref{fig:sibling_ambiguity}. For instance, if the skeleton is traversed in breadth-first search (BFS) order and the preceding skeleton tokens are 1, 2, and 3, the next joint could be either 4 or 5, each being equally valid.

\begin{figure}[h]
\includegraphics[width=0.7\linewidth]{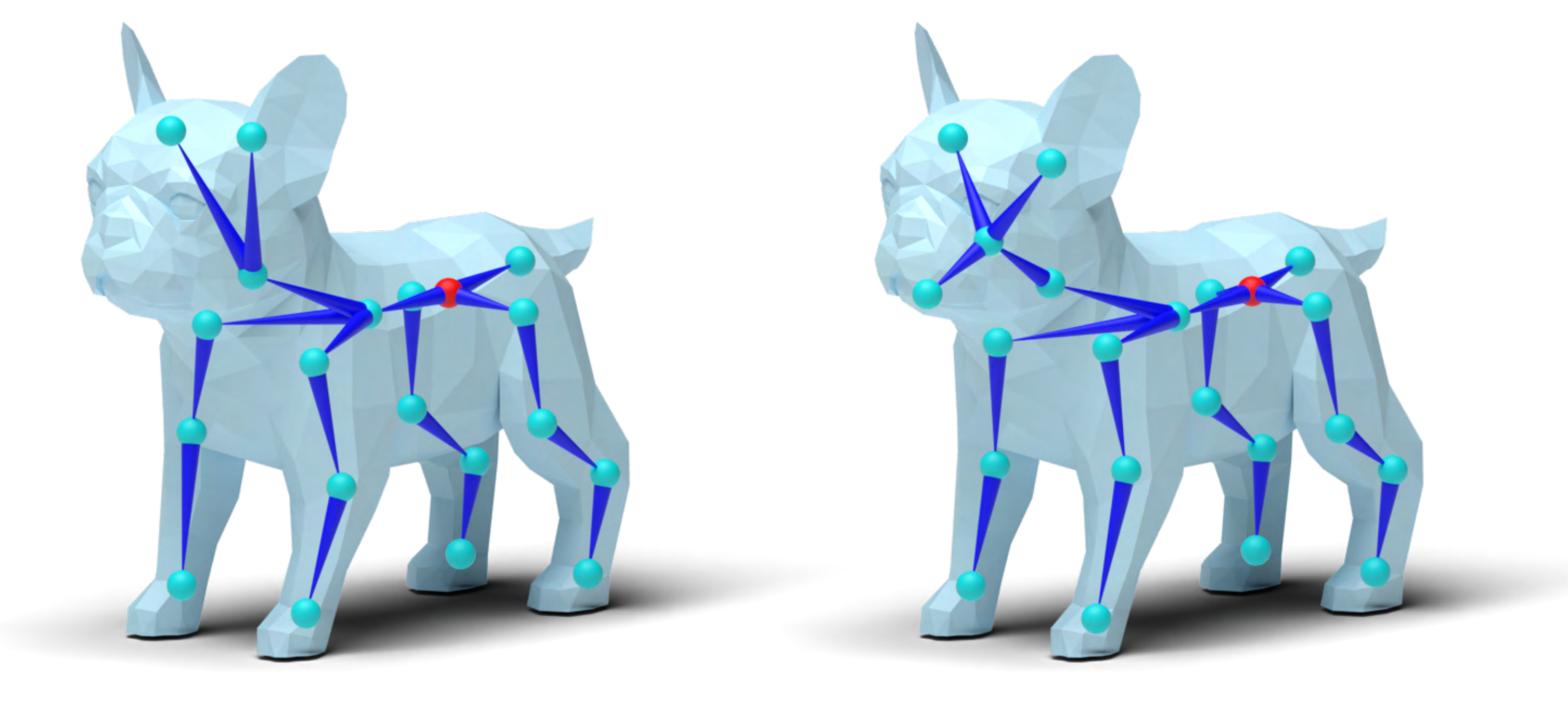}
\captionsetup{labelfont={color=black},textfont={color=black}}
\caption{Examples of different valid skeleton topologies for the same shape.}
\label{fig: diffusion_sampling}
\end{figure}

\textbf{(2) Topology ambiguity:} An object may have multiple valid skeleton topologies, as shown in \cref{fig: diffusion_sampling}. This requires the method to model a distribution over multiple plausible configurations. Our method naturally addresses this ambiguity by modeling the distribution of the next joint based on preceding predictions, which offers a distinct advantage over deterministic approaches in capturing the inherent uncertainty in joint positions.

\subsection{Autoregressive Skeleton Prediction}\label{sec:skeleton}
\boldstartspace{Autoregressive Modeling.} The key component of our method is an autoregressive model for the skeleton prediction to address the ambiguity of skeleton structures and eliminates the need for predefined templates.
To convert the tree-structured skeleton to a sequence that can be effectively processed by the autoregressive model, we adopt the BFS to serialize the skeleton to a list: 

\begin{equation}
    \mathcal{J} = \left[ (j_1, p_1), (j_2, p_2), ..., (j_K, p_K) \right],
\end{equation}

where $j_k \in \mathbb{R}^3$ and $p_k \in \{ 1, ..., K \}$ denote the 3D position and the parent index of the $k$-th joint respectively. As we adopt the BFS order, $p_k<k$ and the first element $(j_1,p_1)$ always represents the root joint. The order of joints at the same BFS depth level is non-deterministic.
To resolve this ambiguity, we randomly sample the order during training and use generative modeling to cover the uncertainty.

Given an input shape $\Shape$ represented by $L$ sampled points, we factorize the joint probability of skeleton by the chain rule:

\begin{align*}
    P(\mathcal{J} \mid \Shape) &= \prod_{k=1}^{K} P\left(j_k, p_k \mid \mathcal{J}_{1:k-1}, \Shape\right).
\end{align*}
where $\mathcal{J}_{1:k}$ is the shorthand for the sublist of $\mathcal{J}$ up to the $k$-th element. 

The autoregressive model is tasked to iteratively predict the conditional distribution of each joint position $j_k$ and parent index $p_k$, formulated as
\begin{align*}
P\left(j_k, p_k \mid \mathcal{J}_{1:k-1}, \Shape\right)
& = P\left(j_k\mid \mathcal{J}_{1:k-1}, \Shape\right) P\left(p_k\mid j_k, \mathcal{J}_{1:k-1}, \Shape\right).
\end{align*}
 
Instead of directly modeling in the original joint space, we map all previously predicted joints and their corresponding parents into a higher-dimensional token space to enhance the model's expressive capacity. This token space effectively represents the evolving state of the skeleton, capturing its structural and hierarchical information as new joints and connections are incrementally added. Similarly, a sequence of shape tokens is extracted to encapsulate the global structure of the input shape, providing consistent contextual information throughout the modeling process. Denoting the skeleton tokens as $T_{1:k-1}\in\mathbb{R}^{(k-1)\times d}$ and the shape token as $H\in\mathbb{R}^{L\times d}$, \revise{where $d$ is the dimension of each token}. The prediction targets are reformulated as:  
\begin{equation}
P(j_k \mid T_{1:k-1}, H) \quad \text{and} \quad P(p_i \mid j_i, T_{1:k-1}, H).
\end{equation}
\revise{The extraction of the skeleton token $T$ and shape token $H$ are detailed in \cref{sec:transformer}.}

\boldstartspace{Joint Prediction with Diffusion Model.} 
To predict the next joint position, which is continuously valued, we address the limitation that most autoregressive models are traditionally designed for discrete outputs, making them less effective for continuous-valued tasks. Inspired by recent autoregressive image generation models~\cite{li2024autoregressive}, we adopt a diffusion sampling process~\cite{ho2020denoising, nichol2021improved, dhariwal2021diffusion} to handle the continuous nature of joint positions. Diffusion models are particularly suited for this task because they iteratively refine samples, effectively resolving the structural ambiguities inherent in skeleton tree representations.
For readability, we drop the current joint index $k$ in the following part.

\paragraph{Forward Diffusion Process:}
The forward process gradually adds Gaussian noise to the ground-truth joint \(j^0\) over \(M\) time steps, producing increasingly noisy versions \(j^m\). This is formulated as:
\[
j^m = \sqrt{\bar{\alpha}_m} j^0 + \sqrt{1 - \bar{\alpha}_m} \epsilon,
\]
where \(\epsilon \sim \mathcal{N}(\mathbf{0}, \mathbf{I})\) is Gaussian noise, and \(\bar{\alpha}_m = \prod_{s=1}^m \alpha_s\) defines a noise schedule.

\paragraph{Training Objective:}
We train a noise estimator \(\epsilon_\theta\), conditioned on the diffusion time step \(m\) and the context \(Z\)$\in\mathbb{R}^{\left(L+k-1\right)\times d}$, where 
\begin{equation}
Z = \text{TransformerBlocks}(T_{1:k-1}, H),
\end{equation}
capturing both the evolving skeleton state and the input shape. 
The model takes the noisy joint \(j^m\) as input and predicts the added noise \(\epsilon\). The training objective is defined as:
\begin{equation}
    \mathcal{L}_{\text{joint}}(Z, j^0) = \mathbb{E}_{\epsilon, m} \big[ \| \epsilon - \epsilon_\theta(j^m \mid m, Z) \|^2 \big].
\end{equation}

\paragraph{Reverse Diffusion Process:}
At inference time, the reverse process iteratively removes noise, sampling the next joint position \(j^0 \sim p_\theta(j^0 \mid Z)\). Starting from a Gaussian sample \(j^M \sim \mathcal{N}(\mathbf{0}, \mathbf{I})\), the reverse process is defined as:
\begin{equation}
j^{m-1} = \frac{1}{\sqrt{\alpha_m}} \big(j^m - \frac{1 - \alpha_m}{\sqrt{1 - \bar{\alpha}_m}} \epsilon_\theta(j^m \mid m, Z)\big) + \sigma_m \delta,
\end{equation}
where \(\delta \sim \mathcal{N}(\mathbf{0}, \mathbf{I})\) is Gaussian noise, and \(\sigma_m\) denotes the noise level at step \(m\). The final output \(j^0\) represents the predicted joint position.

\boldstartspace{Connectivity Prediction.} After we sample the next joint position $j_{k} \in \mathbb{R}^3$ from the diffusion module described earlier, we aim to predict how this newly sampled joint $j_{k}$ connects to its ancestor joints. We first update the context $Z_k$ with the sampled joint $j_k$ through a fusing module F:
\begin{equation}
    Z'_{k} = \text{F} \Bigl ( Z_k, j_{k}, \gamma(k) \Bigr ),
\end{equation}
where $\gamma (k)\in \mathbb{R}^d$ is a positional embedding signaling the current joint index.

Next, a connectivity module C takes $Z'_{k}$ and each individual predicted skeleton token $T_i (i < k)$ (detailed in \cref{sec:transformer})  produce the parent joint probability,   
\begin{equation}
    \mathbf{q}_{k} = \text{Softmax}\Bigl ([ \text{C} (Z'_{k}, T_i) ]_{i=1}^{k-1} \Bigr ).
\end{equation}

The connectivity is supervised with the binary cross-entropy loss,
\begin{equation}
    \mathcal{L}_{\text{connect}} 
    = - \sum_{i=1}^{k-1}
    \bigl[
        \hat{y}_{k,i} \log\bigl(q_{k,i}\bigr)
        \;+\;
        \bigl(1 - \hat{y}_{k,i}\bigr) \log\bigl(1 - q_{k,i}\bigr)
    \bigr],
\end{equation}
where $q_{k,i}$ is the $i$-th element in $\mathbf{q}_k$ and $\hat{y}_{k,i} \in \{0, 1 \}$ is the ground-truth label indicating whether joint $j_{k}$ is connected to joint $j_i$.

During training, the ground-truth next joint position $j_{k}$ is fed into the network for connectivity prediction, while during the inference time, $j_{k}$ is sampled from the joint diffusion module and subsequently passed to the connectivity network.

\begin{figure}[t]
\includegraphics[width=\linewidth]{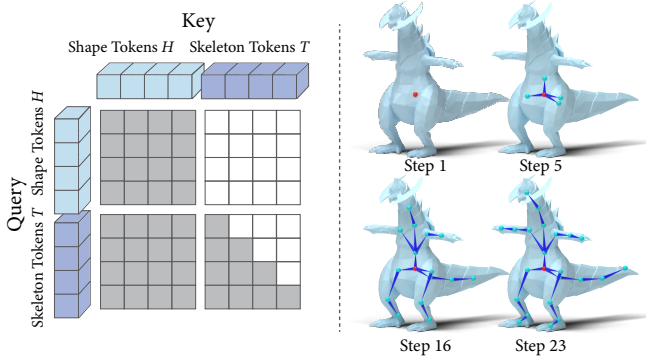}
\caption{(Left) Hybrid attention mask: Shape tokens use full self-attention, while skeleton tokens attend to shape tokens and apply causal masking among themselves. (Right) The skeleton sequence is autoregressively generated during inference.}
\label{fig:autoregressive}
\end{figure}

\begin{figure*}[t]
\includegraphics[width=\linewidth]{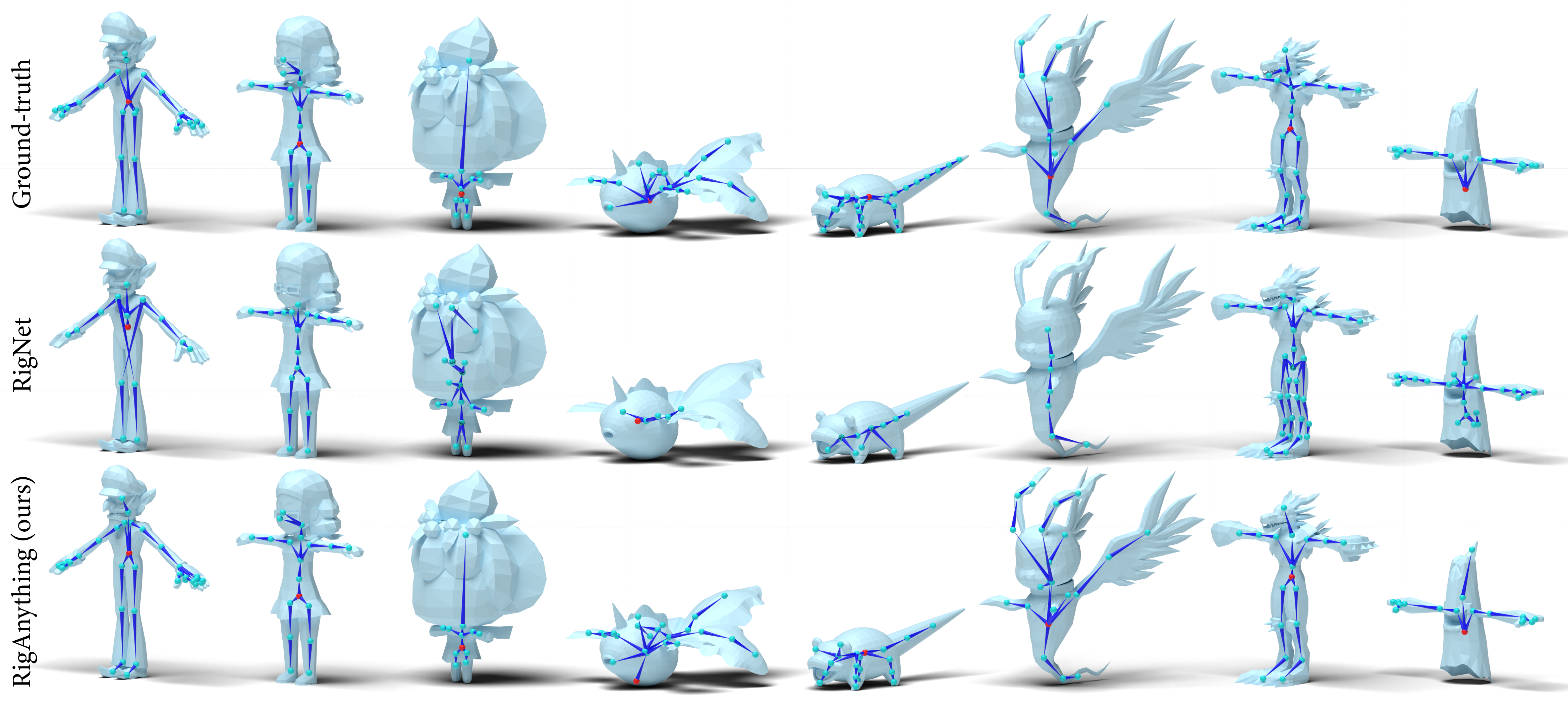}
\caption{Comparison of reconstructed skeletons between our method, RigNet, and ground truth. Our method generates more accurate and satisfying skeletons across diverse shape categories. While RigNet tends to produce excessive joints and struggles with uncommon shapes like characters with tails or wings. Our approach generates a reasonable number of joints and aligns the skeletons closely with the underlying shapes. Note that RigNet supports only rest poses, so all evaluations are conducted on rest-posed objects for fairness.}
\label{fig: skeleton_comp}
\end{figure*}

\subsection{Skinning Prediction} \label{sec:skinning}
Skinning weights are described by a matrix \(W \in \mathbb{R}^{L \times K}\), where each element \(w_{lk}\) indicates the influence of the \(k\)-th joint on the \(l\)-th surface point in \(\mathcal{S}\). The weight vector \(\mathbf{w}_l \in \mathbb{R}^{K}\) for each surface point must satisfy the following constraints:
$\sum_{k=1}^{K} w_{lk} = 1 \quad \text{and} \quad w_{lk} \geq 0 \quad \text{for all } k.$

To compute the skinning weight \revise{\(\mathbf{w}_l\)} for each surface point \(s_l \in \mathcal{S}\), a skinning prediction module G takes as input the shape token $H_{s_l} \in \mathbb{R}^d$ for point $s_l$, along with 
the skeleton token \(T_k\) for each joint \(j_k \, (k \leq K)\). The module outputs a predicted influence score for each joint \(j_k\) on \(s_l\). The final skinning weight \(\mathbf{w}_l\) is computed using the softmax function:
\begin{equation}
\mathbf{w}_l = \text{Softmax}\Bigl([\text{G}(H_{s_l}, T_k)]_{k=1}^{K}\Bigr),    
\end{equation}

We train this module by minimizing a weighted cross-entropy loss, where the ground-truth skinning weight $\hat{\mathbf{w}}_l$ serves as the weighting factor, which can be written as. 
\begin{equation}
    \mathcal{L}_{\text{skinning}}
    = \frac{1}{L} \sum_{l=1}^{L} 
    \Bigl(- \sum_{k=1}^{K} \hat{w}_{l,k} \,\log\bigl(w_{l,k}\bigr)\Bigr).
\end{equation}
This formulation encourages the model to produce higher probabilities for joints with larger ground-truth skinning weights, thereby aligning the learned distribution with the correct influences for each point.

\subsection{Autoregressive Transformer Architecture}\label{sec:transformer}
Our autoregressive modeling is anchored on a transformer-based architecture, which outputs the shape tokens $H\in \mathbb{R}^{L \times d}$ and skeleton tokens $T_{1:k}\in\mathbb{R}^{k\times d}$ ($0< k\leq K$) that serve as conditional inputs for the autoregressive modeling for skeleton prediction (\cref{sec:skeleton}) and skinning prediction (\cref{sec:skinning}).
The extraction of these tokens involve two steps: first, referred as the ``tokenization'' step, an initial shape token and skeleton tokens are lifted from the raw input, this step produces a higher dimensional vector that has sufficient capacity in preparation to capture richer information in the further processing steps in the transformer; subsequently, the transformer process the tokens through a series of attention blocks with carefully crafted attention masking to obtain the final shape and skeleton tokens, which are finally used as inputs to the skeleton and skinning prediction modules in \cref{sec:skeleton} and \cref{sec:skinning}.

\boldstartspace{Tokenization.} 
For the shape data, we sample a set of $L$ surface points $S \in \mathbb{R}^{L \times 3}$ and concatenate them with their corresponding normals $N \in \mathbb{R}^{L \times 3}$, forming a sequence of \revise{$L$} tokens each with $6$ dimensions. 
These tokens are then passed through MLP layers to a $d$-dimensional space. 
Formally, the shape tokens $H \in \mathbb{R}^{L \times d}$ can be written as
\begin{equation}
    H = \text{MLP} \bigl ( \text{Concat} (S, N) \bigr ).
\end{equation}

For the skeleton data, we first apply MLPs to project each joint position $j_k$ and its corresponding parent joint position $j_{p_k}$ into a $d$-dimensional space. These features are then concatenated with positional embeddings, which encode the index of each joint within the sequence. 
Finally, the concatenated features are processed through MLP layers to obtain the per-joint skeleton tokens. These steps can be expressed formally as
\begin{equation}
    T_{k} = \text{MLP}\Bigl( \text{Concat}\bigl( \text{MLP}(j_k),\; \gamma(k),\; \text{MLP}(j_{p_k}),\; \gamma(p_k) \bigr) \Bigr).
\end{equation}
The skeleton token $T_{1:K}\in \mathbb{R}^{K\times d}$ is a sequence of individual per-joint tokens in BFS-order.

\boldstartspace{Processing Tokens with Transformer.} The extracted shape tokens $H$ and predicted skeleton tokens $T_{1:k-1}$ are concatenated and then treated as $L + \left(k-1\right)$ individual tokens. These are then passed through a chain of transformer blocks, in which multi-head self-attention mechanisms ensure that the skeleton tokens and the shape tokens are aware of each other's features, enabling the model to capture rich global information and interdependencies between the shape context and the evolving skeleton structure.
We propose a hybrid attention mechanism that applies different attention patterns to shape and skeleton tokens. As shown in the left part of \cref{fig:autoregressive}, shape tokens attend to each other via full self-attention to capture global geometric context. 
For skeleton tokens, we first allow them to attend to all shape tokens to incorporate shape information, and then apply causal attention~\cite{waswani2017attention,radford2019language} among the skeleton tokens so that each token only attends to its preceding tokens in the sequence. This ensures the autoregressive property required for skeleton sequential generation. 

The output of the last transformer block $Z_k$ will be served as the condition in the diffusion model for joint $j_k$ sampling as introduced in \cref{sec:skeleton}.

\subsection{Final Training Objective} \label{sec:final_loss}
We train our entire model end-to-end, ensuring that joint positions, connectivity, and skinning weights are learned in a mutually reinforcing manner. Specifically, we combine the losses from the joint diffusion, connectivity, and skinning prediction modules into a single objective. The integrated objective allows the network to learn coherent skeleton structures and accurate skinning assignments simultaneously.

\begin{equation}
    \mathcal{L} = \mathcal{L}_{\text{joint}} + \mathcal{L}_{\text{connect}} + \mathcal{L}_{\text{skinning}}.
\end{equation}

%% file: Sections/4_experiments.tex
\section{Experiments}

\subsection{Implementation Details}

Our input point cloud consists of 1024 points, with the maximum number of joints per sample set to 64. The point cloud and joint tokenizers are implemented as two-layer MLPs with hidden dimensions of 512 and 1024. For both parent and skinning prediction modules, we employ two-layer MLPs with hidden dimensions of 1024.

The implemented transformer consists of 12 layers with a hidden dimension of 1024. Following the implementation in \cite{zhang2025gs}, each transformer block incorporates a multi-head self-attention layer with 16 heads and a two-layered MLP with a hidden dimension of 4096 and a GeLU activation. We employ Pre-Layer Normalization, Layer Normalization (LN), and residual connections consistent with the reference implementation. During training, we employ a hybrid attention masking strategy: shape tokens perform self-attention to effectively capture geometric information, while skeleton tokens use causal attention, attending only to their ancestor skeleton tokens within the sequence to facilitate auto-regressive generation. Additionally, skeleton tokens attend to all shape tokens. During inference, the network processes shape tokens as input and generates skeleton tokens in an auto-regressive manner. \revise{We apply KV caching in the transformer blocks to boost the inference speed.}

The joint diffusion process follows \cite{nichol2021improved, li2024autoregressive}, which has a cosine noise scheduler with 1000 training steps and \revise{50} resampling steps during inference. The denoising MLP is conditioned on the transformer-outputted joint tokens, where these tokens are incorporated into the noise scheduler's time embedding through AdaLN \cite{peebles2023scalable} within the Layer Normalization layers.

The fusing module is a two-layer MLP with an input size of 3072 and hidden dimensions of 2048 and 1024. During inference, after obtaining the next joint position via diffusion sampling, a shape tokenizer generates a latent shape token (dimension 1024), which is concatenated with previous context tokens (dimension 1024) and positional embeddings. The fusing module's output serves as the updated context for connectivity and skinning prediction. Both the connectivity and skinning modules share a similar architecture with the fusing module, except that their input size is 2048.

In the autoregressive skeleton generation process, we maintain a learnable start token (BOS) at the beginning of each skeleton sequence to indicate the start of generation. As we traverse the skeleton in BFS order, the joint sampled from the start token is considered the root joint. The stop condition is determined by checking the parent of the current joint—if a joint's parent is itself, it indicates that the sequence has reached its end.

\begin{figure}[t]
\includegraphics[width=\linewidth]{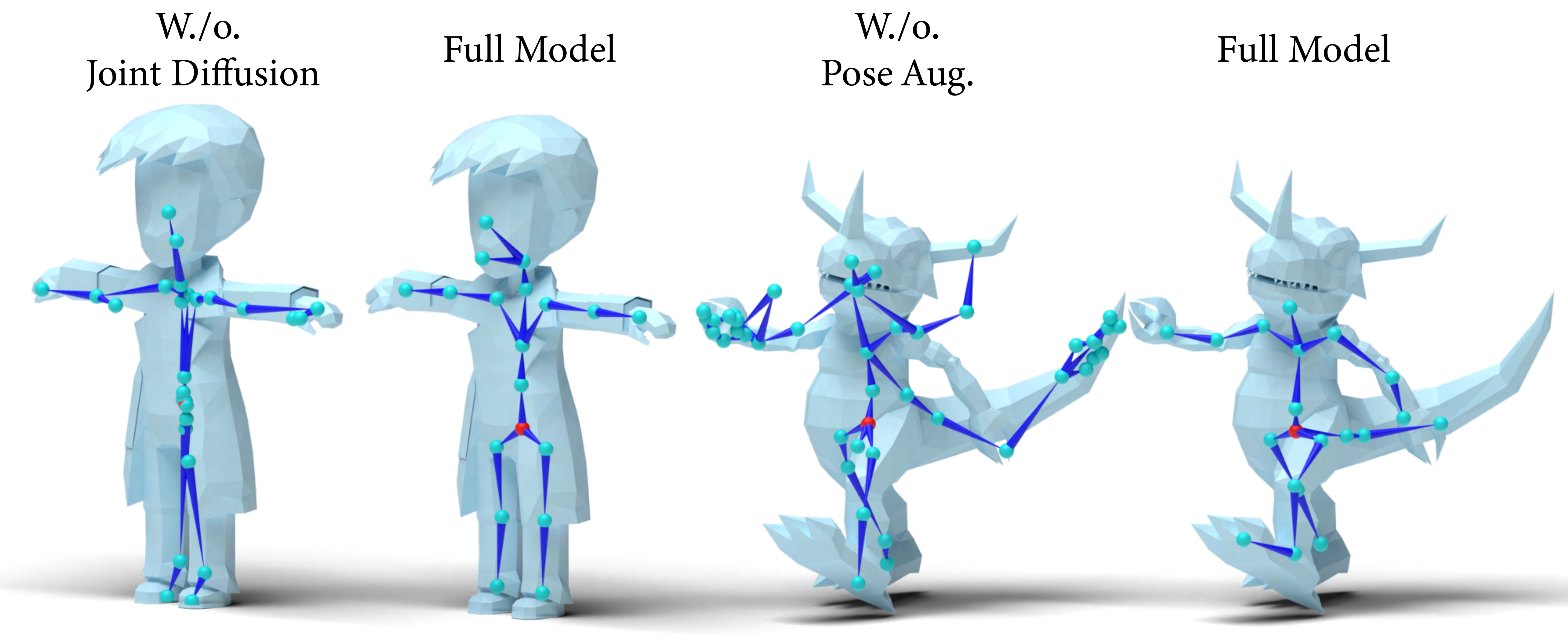}
\caption{(Left) Joint diffusion modeling prevents joint collapse to mean positions, capturing diverse modalities. (Right) Pose augmentation improves generalization to unseen poses, ensuring well-aligned skeletons and avoiding excessive joints.}
\label{fig:ablation_diffusion}
\end{figure}

\subsection{Dataset}

\boldstartspace{Overview.} We utilize both the RigNet dataset \cite{rignet} and the Objaverse dataset \cite{deitke2023objaverse}. The RigNet dataset contains 2,354 high-quality 3D models with ground-truth skeleton and skinning annotations. The Objaverse dataset offers a large collection of 3D models with varying rigging quality. To ensure data reliability, we filtered out 9,686 models with consistent and accurate skeleton and skinning information. Our dataset spans a diverse range of categories, including bipedal, quadrupedal, avian, marine, insectoid, and manipulable rigid objects. For each model, we sample point clouds and face normals from the mesh surface.

During training, we perform online pose augmentation to the input data by randomly deforming the input point clouds using the ground-truth skeleton and skinning. \revise{Specifically, we apply random perturbations to each joint, with the maximum rotation angle constrained to 45 degrees, and then deform the original point cloud using the perturbed skeleton and the ground truth skinning.} As shown in our ablation study in \cref{sec:ablation} and \cref{fig:ablation_diffusion}, this augmentation strengthens our method's ability to generalize to objects in different poses.

\boldstartspace{Data Filtering} We apply a thorough data filtering process to ensure the quality and validity of the dataset. Both manual checks and automated scripts are used to maintain consistency and quality. The filtering is based on the following rules:

\begin{itemize}
    \item Shapes with more than 64 joints are excluded.
    \item Shapes with invalid skeletons (e.g., skeleton hierarchy not forming a proper tree) are excluded.
    \item Shapes whose skeletons do not align well with the geometry are excluded.
    \item Overly simplified or indistinct shapes (e.g., consisting of very few vertices and faces) are excluded.
\end{itemize}

The Objaverse dataset originally contains 21,622 shapes with rigging annotations. We exclude 811 shapes with overly complex skeletons (more than 64 joints), which are often related to facial or hair rigs. Additionally, 10,471 shapes are removed due to low-quality rigging or skinning annotations. After filtering, we obtain a refined set of 12,040 shapes with reliable rigging information.

\begin{figure}[h]
\includegraphics[width=0.8\linewidth]{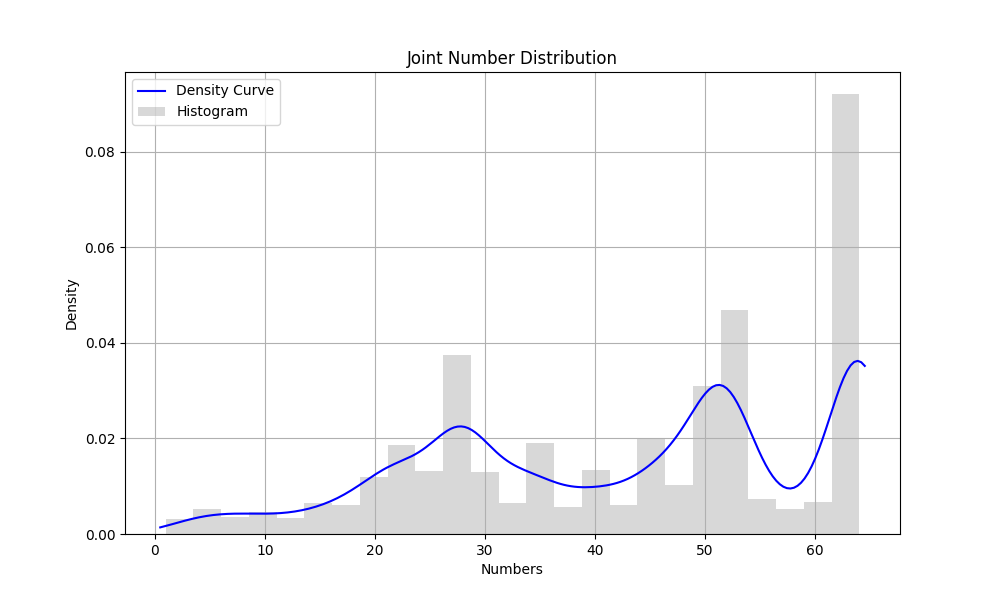}
\caption{Distribution of joint numbers across shapes in our dataset.}
\label{fig:joint_number_distribution}
\end{figure}

\boldstartspace{Data Statistics} We organize the shapes in the Objaverse dataset into six categories based on their labels: humanoid/bipedal, quadruped, insectoid, avian, marine, and other. The "other" category mainly includes manipulable articulated rigid objects, such as suitcases, cabinets, and similar items.
The number of shapes in each category is summarized in \cref{tab:dataset_stat}. We also analyze the distribution of joint counts per shape and present the results in \cref{fig:joint_number_distribution}. Our dataset contains more shapes with joint counts falling in the ranges $[25, 30]$, $[50, 55]$, and $[60, 64]$.

\begin{table}[h]
    \label{tab:dataset_stat}
    \small
    \centering
    \setlength{\tabcolsep}{5pt}
    \scalebox{0.92}{
    {\fontsize{9pt}{10pt}\selectfont
    \begin{tabular}{@{}lc@{}}
    \toprule
        & Number\\ \midrule
    Humanoid/Bipedal  & 7459\\
    Quadruped  &  543 \\
    Insectoid & 129 \\
    Avian & 176 \\
    Marine & 251 \\
    Other & 830 \\ \midrule
    \textbf{Total} & 9388 \\
    \bottomrule
    \end{tabular}}}
    \caption{Category statistics of the filtered Objaverse dataset.}
\end{table}

\subsection{Evaluation and Baseline Comparisons}

\begin{figure*}[t]
\includegraphics[width=0.8\linewidth]{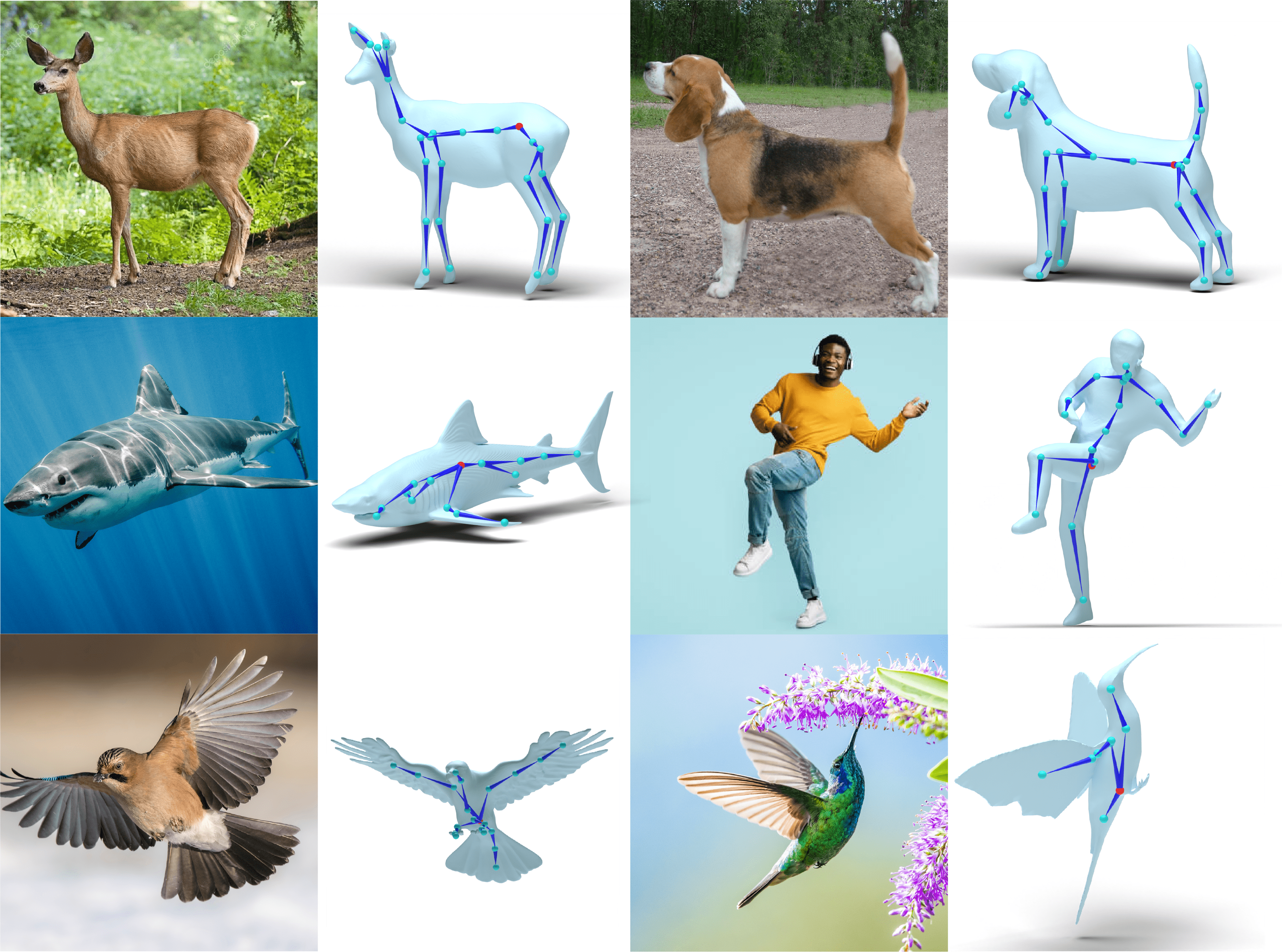}
\caption{Skeleton results on shapes from real casual images. We use off-the-shelf image-to-3D model pipeline \revise{\cite{liu2024meshformer}} to generate the shapes from real images and apply RigAnything to predict their skeletons.}
\label{fig:real_capture}
\end{figure*}

\subsubsection{Skeleton Prediction}

\begin{figure*}[t]
\includegraphics[width=\linewidth]{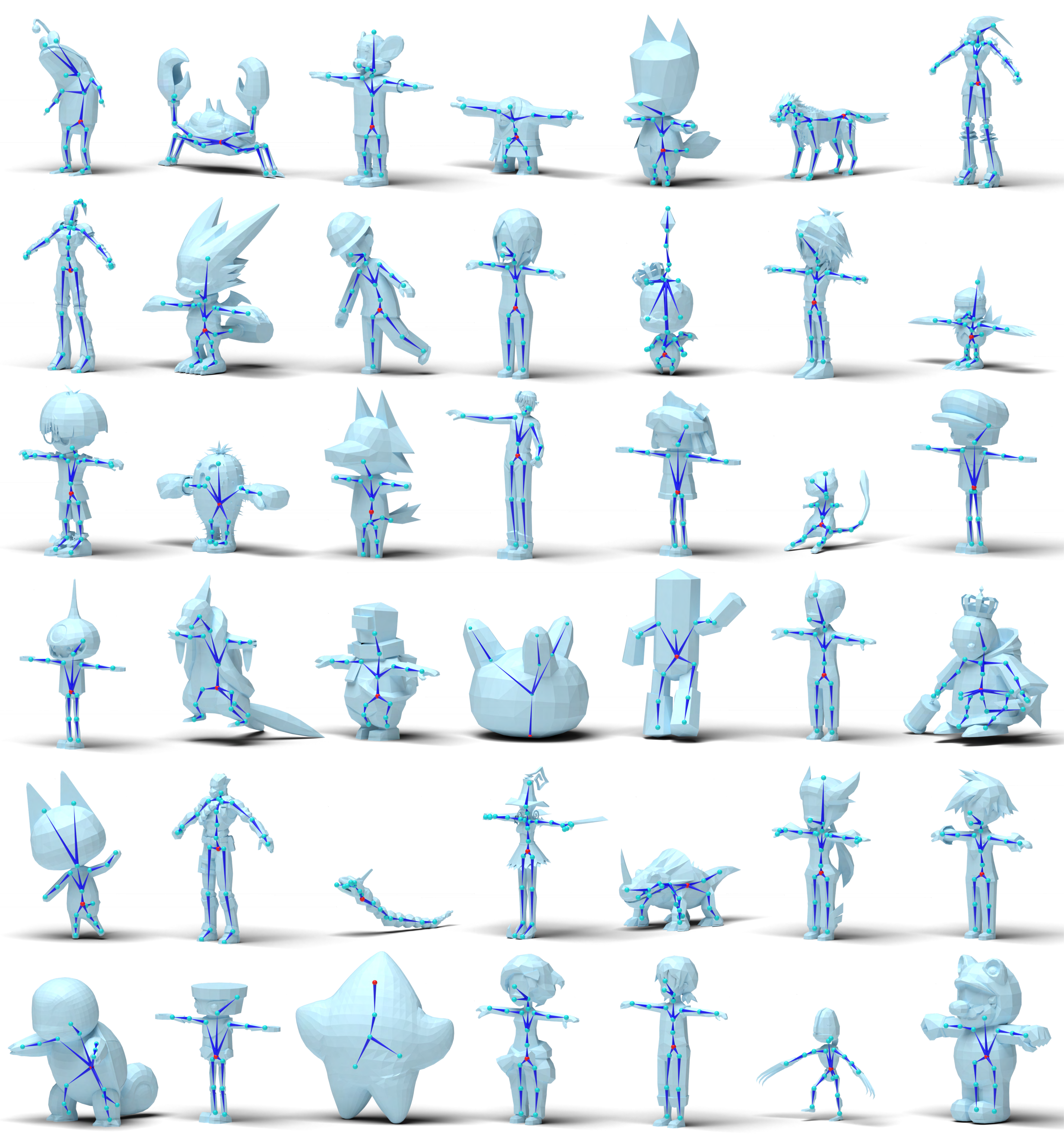}
\caption{More results on the RigNet dataset. Please refer to the supplementary video for more 360-degree video results on both the RigNet and Objaverse dataset.}
\label{fig: moreresults}
\end{figure*}

We provide qualitative visualizations of the reconstructed skeletons in comparison with the ground truth and RigNet in \cref{fig: skeleton_comp}. Our method demonstrates superior performance, producing more accurate and satisfying skeletons across various shape categories. In contrast, RigNet~\cite{rignet} struggles to recover reasonable skeletons for less common shapes, such as characters with tails or wings, and frequently generates an excessive number of joints. In comparison, our method generates a reasonable number of joints, with the reconstructed skeletons well-aligned to the underlying shape, ensuring better structural consistency and fidelity. \revise{More skeleton prediction results of our method can be found in \cref{fig: moreresults}.}

\input{Tabels/skel_prediction}

To quantitatively evaluate the performance of skeleton prediction, we measure the similarity between the predicted skeletons and the ground truth using \revise{metrics proposed by RigNet~\cite{rignet}}: Intersection over Union (IoU), Precision, and Recall for bone matching, as well as Chamfer distances for joints (CD-J2J), bone line segments (CD-B2B), and joint-to-bone line segments (CD-J2B). \revise{We train and test both our method and the baseline on the same RigNet + Objaverse dataset. Table \ref{tab:skel_pred} presents a comparison with RigNet~\cite{rignet} across these metrics. Note that we revised the original RigNet implementation to accommodate non-symmetric models.} The results show that our method significantly outperforms the baselines, producing skeletons that align more closely with GT.

\revise{We compare our method with humanoid auto-rigging approaches, including TARig~\cite{ma2023tarig} and Neural Blend Shapes (NBS)~\cite{li2021learning}, and present the results in Table \ref{tab:skel_pred_humanoid}. All methods are evaluated on the humanoid subset of the original evaluation sets. The results show that our method produces significantly better skeletons. Note that TARig and NBS cannot be finetuned, as TARig does not provide training scripts and NBS did not release data processing scripts.}

\input{Tabels/skel_prediction_humanoid}

\subsubsection{Connectivity Prediction}

We evaluate the connectivity prediction performance when the given joints are from ground truth instead of prediction. We measure the binary classification accuracy (Class. Acc.) for assessing joint pair connections, as well as the CD-B2B and edit distance (ED), which measure the geometric and topological difference between the predicted and reference skeletons. As shown in Table \ref{tab:connect_pred}, our method significantly outperforms RigNet across all metrics.

\input{Tabels/connect_pred}

\subsubsection{Skinning Prediction}

\begin{figure}[t]
    \includegraphics[width=\linewidth]{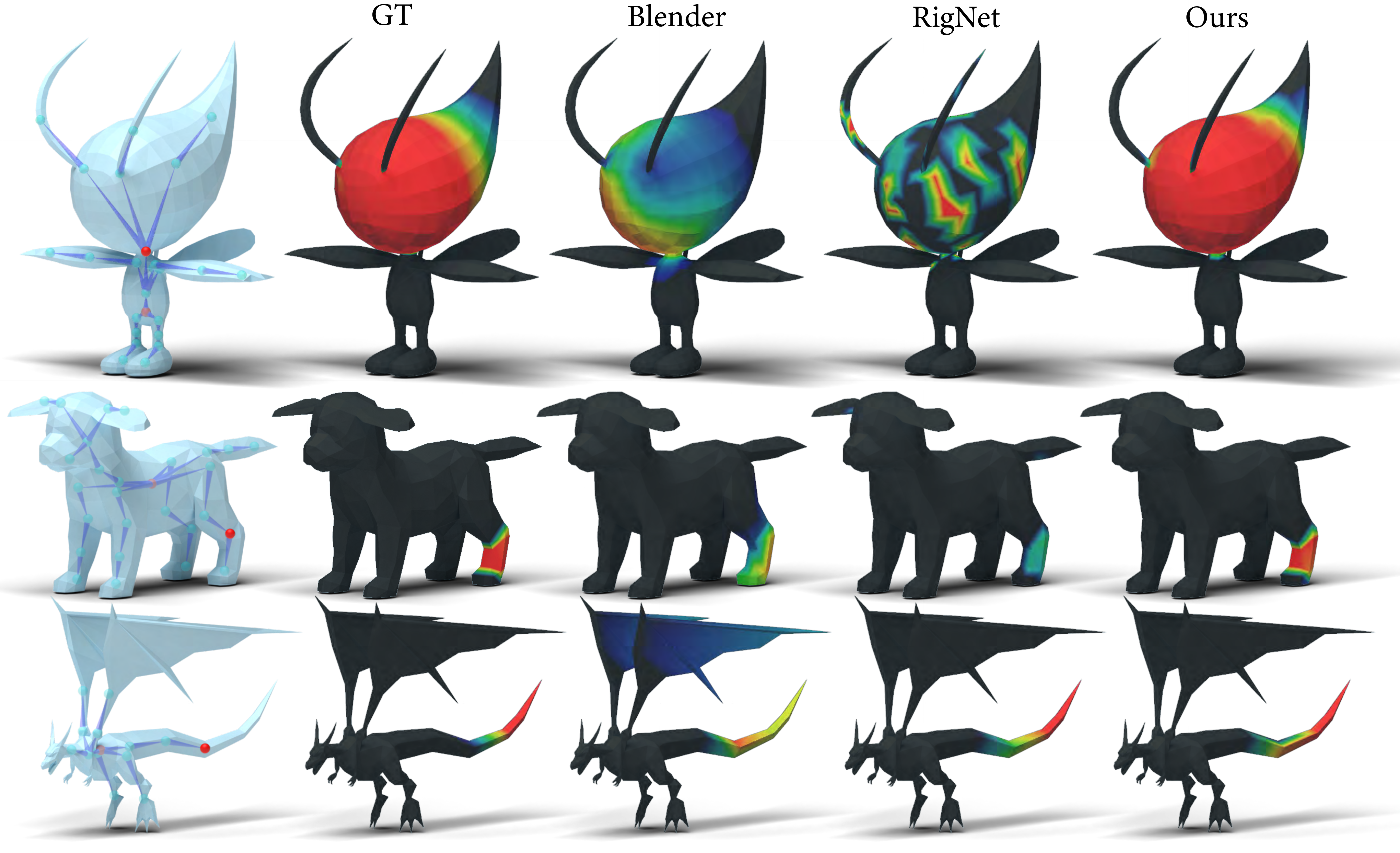}
    \caption{Comparison of skinning weight predictions. Our method produces more accurate and consistent weights, especially in challenging cases with large geodesic distances.}
    \label{fig:skin_pred}
\end{figure}

\revise{
We train both the baseline and our method on the new RigAnything dataset and quantitatively evaluate skinning weight prediction using the metrics proposed in the original RigNet and its evaluation set, with results shown in Table \ref{tab:skin_pred}. Our method achieves higher precision and lower average L1 error without relying on geometric priors. 
}
We also qualitatively compare our method with RigNet and Blender’s built-in automatic skinning, which assigns weights based on the shortest Euclidean distance to bones. For fairness, we use the ground truth skeleton during inference. As shown in \cref{fig:skin_pred}, our method produces more accurate and consistent weights, especially when differentiating regions that are close in Euclidean space but far in geodesic distance, where baselines fail.

\input{Tabels/skin_pred}

\subsection{Ablation Study} \label{sec:ablation}

We analyze various components of our method and compare their performance with the final model.

\subsubsection{Joint Diffusion} In our full model, the joint diffusion module predicts the probability of the next joint position based on preceding joints in a skeleton sequence. This probabilistic approach effectively resolves structural ambiguities in skeleton tree representations, such as equivalent sibling node orderings, by accounting for their equivalence. In an ablation study, we replaced the joint diffusion loss with a deterministic L2 joint position loss. As shown in \cref{fig:ablation_diffusion}, using L2 loss leads to joints collapsing toward the middle axis, representing the mean position across samples due to sibling ambiguities within the skeleton sequence. In contrast, our method captures diverse joint position modalities, producing reasonable and accurate joint placements instead of averaged positions. Quantitative results in \cref{tab:ablation_skel_pred} further confirm that joint diffusion modeling significantly improves our method’s performance, boosting the skeleton IoU by almost two times.

\input{Tabels/ablation_skeleton_pred}

\subsubsection{Normal Injecting} To evaluate the impact of incorporating point normals into the shape tokens, we conducted a comparison experiment without point normals as input. The numerical results in \cref{tab:ablation_skel_pred} show a significant decline in skeleton performance when normal information is excluded, highlighting the importance of point normals as geometric information for improving performance.
\revise{We further studied the effect of normal injection on the performance of skinning prediction and provide the results in Table \ref{tab:skin_pred_ablation}. The results demonstrate that normal information improves geodesic inference and skinning prediction, as surfaces with similar normals tend to be geodesically close. This helps the network infer connectivity information that is otherwise missing in point cloud representations compared to meshes.}

\input{Tabels/skin_pred_ablation}

\subsubsection{Online Pose Augmentation} We analyze the effect of online data augmentation by randomly deforming input point clouds using the ground-truth skeleton and skinning. As shown in the numerical results in \cref{sec:ablation}, pose augmentation improves skeleton prediction performance. Additionally, \cref{fig:ablation_diffusion} compares results with and without pose augmentation on a character with a random skeleton pose not present in the dataset. Our full model generates a significantly better-aligned skeleton structure, whereas the model trained without pose augmentation fails to produce skeletons aligned with the shape and generates excessive joints. This augmentation enhances our method's ability to generalize to objects in diverse poses. Furthermore, as demonstrated in \cref{fig:real_capture}, our method achieves high-quality skeletons even when the input shapes are obtained from real-world data and the targets are in arbitrary poses.

%% file: Tabels/skel_prediction.tex
\begin{table}[H]
    \color{black}
    \small
    \centering
    \setlength{\tabcolsep}{2pt}
    \scalebox{0.92}{
    {\fontsize{9pt}{10pt}\selectfont
    \begin{tabular}{@{}ccccccc@{}}
    \toprule
                   & IoU  $\uparrow$ & Prec. $\uparrow$ & Rec. $\uparrow$ & CD-J2J $\downarrow$ & CD-J2B $\downarrow$ & CD-B2B $\downarrow$ \\ \midrule
    RigNet  & 0.456 & 0.424 & 0.591 & 0.048 &  	0.042 & 0.030 \\
    \textbf{RigAnything}  & \textbf{0.768} & \textbf{0.789} & \textbf{0.766} & \textbf{0.034} & \textbf{0.035} & \textbf{0.020} \\ \bottomrule
    \end{tabular}}}
    \caption{Quantitative comparison of skeleton prediction on \revise{the RigNet + Objaverse dataset}. Our predicted skeletons alignes better with the ground truth.}
    \label{tab:skel_pred}
\end{table}

%% file: Tabels/skel_prediction_humanoid.tex
\begin{table}[H]
    \color{black}
    \small
    \centering
    \setlength{\tabcolsep}{2pt}
    \scalebox{0.92}{
    {\fontsize{9pt}{10pt}\selectfont
    \begin{tabular}{@{}ccccccc@{}}
    \toprule
                   & IoU  $\uparrow$ & Prec. $\uparrow$ & Rec. $\uparrow$ & CD-J2J $\downarrow$ & CD-J2B $\downarrow$ & CD-B2B $\downarrow$ \\ \midrule
    NBS  & 0.337 & 0.313 & 0.377 & 0.124 & 0.107 & 0.377 \\
    TARig  & 0.603 & 0.591 & 0.637 & 0.100 & 0.090 & 0.079 \\
    \textbf{RigAnything}  & \textbf{0.886} & \textbf{0.904} & \textbf{0.884} & \textbf{0.030} & \textbf{0.033} & \textbf{0.018} \\ \bottomrule
    \end{tabular}}}
    \caption{Comparison of skeleton prediction on the humanoid subset. Our method significantly outperforms other humanoid auto-rigging approaches.}
    \label{tab:skel_pred_humanoid}
\end{table}

%% file: Tabels/connect_pred.tex
\begin{table}[H]
    \color{black}
    \label{tab:connect_pred}
    \small
    \centering
    \setlength{\tabcolsep}{5pt}
    \scalebox{0.96}{
    {\fontsize{9pt}{10pt}\selectfont
    \begin{tabular}{@{}cccc@{}}
    \toprule
                   & Class. Acc.  $\uparrow$ & CD-B2B $\downarrow$ & ED $\downarrow$ \\ \midrule
    RigNet  & 0.358 & 0.017 & 12.663 \\
    \textbf{RigAnything}  & \textbf{0.965} & \textbf{0.001}&  \textbf{2.150} \\ \bottomrule
    \end{tabular}}}
    \caption{Connectivity prediction on \revise{the RigNet + Objaverse dataset}. Our method significantly outperforms RigNet across all metrics.}
    \label{tab:connect_pred}
\end{table}

%% file: Tabels/skin_pred.tex
\begin{table}[H]
    \captionsetup{labelfont={color=black},textfont={color=black}}
    \color{black}
    \small
    \centering
    \setlength{\tabcolsep}{5pt}
    \scalebox{0.92}{
    {\fontsize{9pt}{10pt}\selectfont
    \begin{tabular}{cccc}
    \toprule
        & Prec.  $\uparrow$ & Rec. $\uparrow$ & Avg. L1 $\downarrow$ \\ \midrule
    RigNet  &  0.755 & \textbf{0.828} & 0.485 \\
    \textbf{RigAnything} & \textbf{0.825} & 0.798&  \textbf{0.432}\\ \bottomrule
    \end{tabular}}}
    \caption{Quantitative evaluation of skinning weights prediction.}
    \label{tab:skin_pred}
\end{table}

%% file: Tabels/ablation_skeleton_pred.tex
\begin{table}[H]
    \small
    \centering
    \setlength{\tabcolsep}{2pt}
    \scalebox{0.80}{
    {\fontsize{9pt}{10pt}\selectfont
    \begin{tabular}{@{}ccccccc@{}}
    \toprule
                   & IoU  $\uparrow$ & Prec. $\uparrow$ & Rec. $\uparrow$ & CD-J2J $\downarrow$ & CD-J2B $\downarrow$ & CD-B2B $\downarrow$ \\ \midrule
    Ours w./o. joint diffusion & 0.308  & 0.277 & 0.364 & 0.068 & 0.059 & 0.046 \\
    Ours w./o. injecting normal & 0.559 & 0.603 & 0.547 & 0.053 & 0.048 & 0.034 \\
    Ours w./o. pose aug. & 0.741 & 0.768 & 0.732 & 0.037 & 0.037 & 0.022 \\
    Ours full model & \textbf{0.765} & \textbf{0.786} & \textbf{0.765} & \textbf{0.033} & \textbf{0.034} & \textbf{0.019} \\ \bottomrule 
    \end{tabular}}}
    \caption{Ablation study results showing the impact of joint diffusion, normal injection, and pose augmentation on skeleton prediction.}
    \label{tab:ablation_skel_pred}
\end{table}

%% file: Tabels/skin_pred_ablation.tex
\begin{table}[H]
    \captionsetup{labelfont={color=black},textfont={color=black}}
    \color{black}
    \small
    \centering
    \setlength{\tabcolsep}{5pt}
    \scalebox{0.92}{
    {\fontsize{9pt}{10pt}\selectfont
    \begin{tabular}{@{}cccc@{}}
    \toprule
                   & Prec.  $\uparrow$ & Rec. $\uparrow$ & Avg. L1 $\downarrow$ \\ \midrule
    W./o. injecting normal 	  & 0.830 & \textbf{0.841} & 0.413 \\
    W./ injecting normal  & \textbf{0.836} & 0.821&  \textbf{0.397} \\ \bottomrule
    \end{tabular}}}
    \caption{Skinning prediction performance with and without normal injection, indicating normal information improves geodesic inference and skinning prediction.}
    \label{tab:skin_pred_ablation}
\end{table}

%% file: Sections/5_discussion.tex
\section{Limitation and Future Work}

Although our method can automatically rig a variety of objects, there are several limitations and potential avenues for future work.
 First, our current approach does not allow for control over the level of detail in the rigs. Specifically, artists may need varying levels of detail in different parts of an object to achieve different degrees of motion control.
To get this feature, more detailed rigging data shall be collected to allow for finer rigging (\textit{e.g.}, head and hand area), and we can introduce a condition in the network to provide control over the level of detail.
In addition, our method relies solely on geometry information to infer rigs, which can sometimes lack sufficient cues for the rig structure, leading to ambiguities. To improve this, texture information can be incorporated as an additional cue for automatic rigging in future iterations.
Furthermore, our skinning weight prediction does not account for different motion styles, such as those influenced by materials. This limitation could be alleviated by incorporating dynamic data into the training process. However, high-quality dynamic data are scarce, and it would be interesting to explore the possibility of collecting such data to further improve rigging performance.

%% file: Sections/6_conclusion_future_works.tex
\section{Conclusion}

In this work, we present RigAnything, an autoregressive transformer-based method that automatically predicts rigs for 3D assets. 
To address the challenges posed by objects with diverse topologies and eliminate the inherent ambiguities in rigging, RigAnything probabilistically predicts the skeletons and assigns skinning weights, eliminating the need for any templates.
This approach allows RigAnything to be trained end-to-end on both RigNet and the diverse Objaverse dataset, ensuring its versatility.
Extensive experiments highlight the superiority of RigAnything across a wide range of object categories, showcasing its effectiveness and generalizability.

\begin{acks}
    We thank Miyuki Richardson for helping create teaser animation. This work was done when Isabella Liu was a research intern at Adobe Research. Prof. Xiaolong Wang was supported in part by gifts from Amazon and Meta. This work was also partially supported by Hillbot Inc., with Prof. Hao Su officially serving as its CTO.
\end{acks}